\documentclass[runningheads]{llncs}

\usepackage{amsfonts}
\usepackage{hyperref}
\usepackage{xurl}
\usepackage[pdftex]{graphicx}
\usepackage{amsmath}
\usepackage{rotating}

\usepackage{geometry}
\usepackage{color}

\usepackage{soul}

\usepackage[]{multirow}
\usepackage{tabulary}
\usepackage{longtable,array,ragged2e}
\newcolumntype{L}[1]{>{\raggedright\let\newline\\\arraybackslash\hspace{0pt}}m{#1}}
\newcolumntype{C}[1]{>{\centering\let\newline\\\arraybackslash\hspace{0pt}}m{#1}}
\newcolumntype{R}[1]{>{\raggedleft\let\newline\\\arraybackslash\hspace{0pt}}m{#1}}
\usepackage{booktabs,dcolumn,caption}

\begin{document}

\title{A Bibliographic View on Constrained Clustering}

\author{Ludmila I. Kuncheva \and
Francis J. Williams \and
Samuel L. Hennessey}
\authorrunning{L Kuncheva et al.}

\institute{School of Computer Science and Electronic Engineering\\
Bangor University, Bangor, Gwynedd\\
LL57 1UT, United Kingdom
\email{l.kuncheva@bangor.ac.uk}}
\maketitle              

\begin{abstract}
A keyword search on constrained clustering on Web-of-Science returned just under 3,000 documents. We ran automatic analyses of those, and compiled our own bibliography of 183 papers which we analysed in more detail based on their topic and experimental study, if any. This paper presents general trends of the area and its sub-topics by Pareto analysis, using citation count and year of publication. We list available software and analyse the experimental sections of our reference collection. We found a notable lack of large comparison experiments. Among the topics we reviewed, applications studies were most abundant recently, alongside deep learning, active learning and ensemble learning. 

\end{abstract}

\section{Introduction}
Constrained clustering is a topic in semi-supervised machine learning. The success of a clustering result is often judged by an external criterion, be it user satisfaction or a match with a predefined structure or class labels. Constrained clustering is aimed as improving the quality of the resultant partition. Pairwise constraints are most widely used. Must Link (ML) constraints between points $A$ and $B$ state that $A$ and $B$ must be in the same cluster, while Cannot Link (CL) constraints state that $A$ and $B$ must not be in the same cluster. 

Constraints can be derived from a labelled data set, introduced manually by a user, or may come from a real life unsupervised problem. Consider the following example. We are interested in recognising individual animals from a video feed. At the start, there is no information about each individual animal. However, the total number of identities can be assumed within a small range. This will define the number of clusters. Suppose that bounding boxes with all animals in the video have been extracted, and features representations have been created thereof. Clustering the data may give us the animal identities. However, we have further information that can be included to improve the clustering. First, animals which are in the same frame of the video cannot be the same identity. This generates CL constraints. Additionally, tracking software may provide trajectories across frames for the same animal. This will generate ML constraints.  

There are many more ways to introduce constraints in clustering, e.g., cluster size, diameter, cardinality, density, feature-specific logical expressions, and so on. 

The available surveys on constrained clustering are a few years old now~\cite{Dinler16}, \cite{Davidson07}, \cite{Gancarski20}, \cite{Basu08}, focus on a specific topic~\cite{dong2020survey}, or only marginally cover constrained clustering~\cite{van2020survey}.

This paper  presents a snapshot of the state-of-the-art in constrained clustering. We acknowledge that scientific impact may not be faithfully represented by citation count~\cite{Bornmann08}, \cite{Tahamtan19}. However, in the absence of a better metric, we provide illustrative bibliometrics and Pareto analysis. While previous surveys report the details of the works they cover, here we look to identify tendencies, progress, prevalence, and presence of the different themes and topics. 

The rest of the paper is organised as follows. Section~\ref{bibmet} gives a general bibliographic metric of the area. In Section~\ref{sec:topics} we create a rough taxonomy of the constrained clustering topics throughout the literature. Section~\ref{sec:par} shows a Pareto front analysis of the constrained clustering literature and its topics. Section~\ref{sec:sof} displays a list of some available software used for constrained clustering. Section~\ref{sec:exp} analyses and summarises the experimental studies carried out in the literature.

\section{General bibliographic remarks}
\label{bibmet}

\begin{figure}[htb]
    \centering
\begin{tabular}{ccc}
\includegraphics[width = 0.31\textwidth]{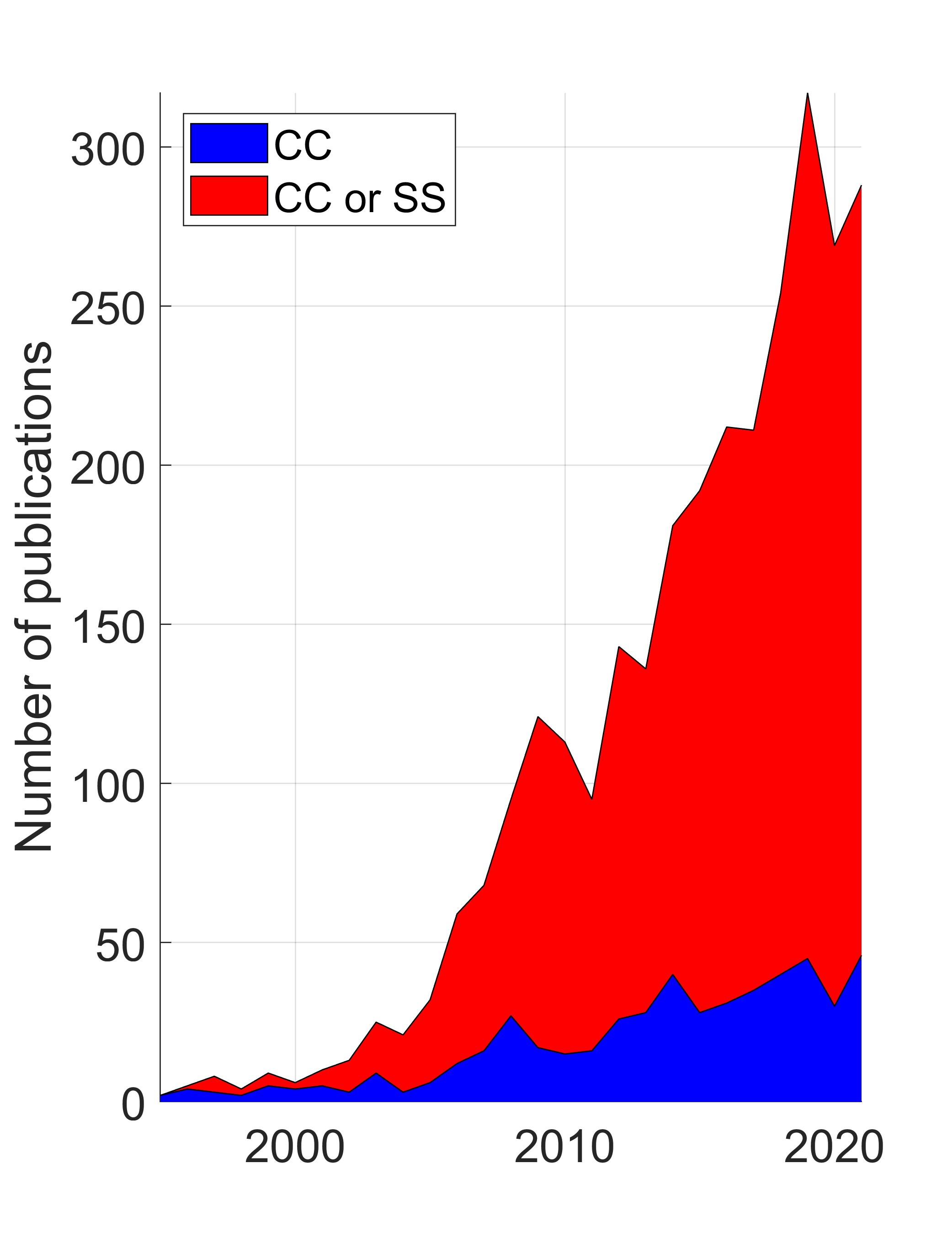}&
\includegraphics[width = 0.31\textwidth]{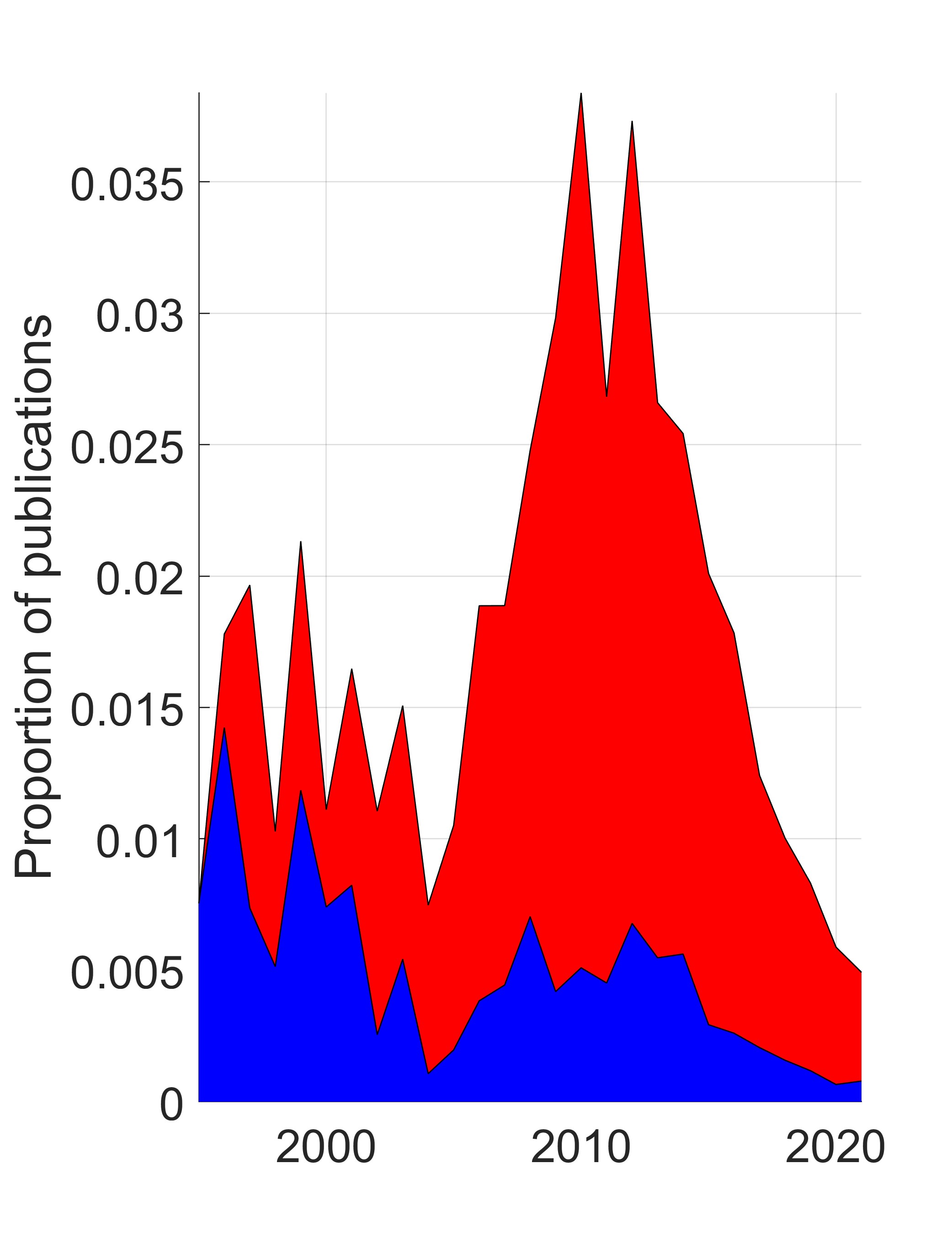}&
\includegraphics[width = 0.31\textwidth]{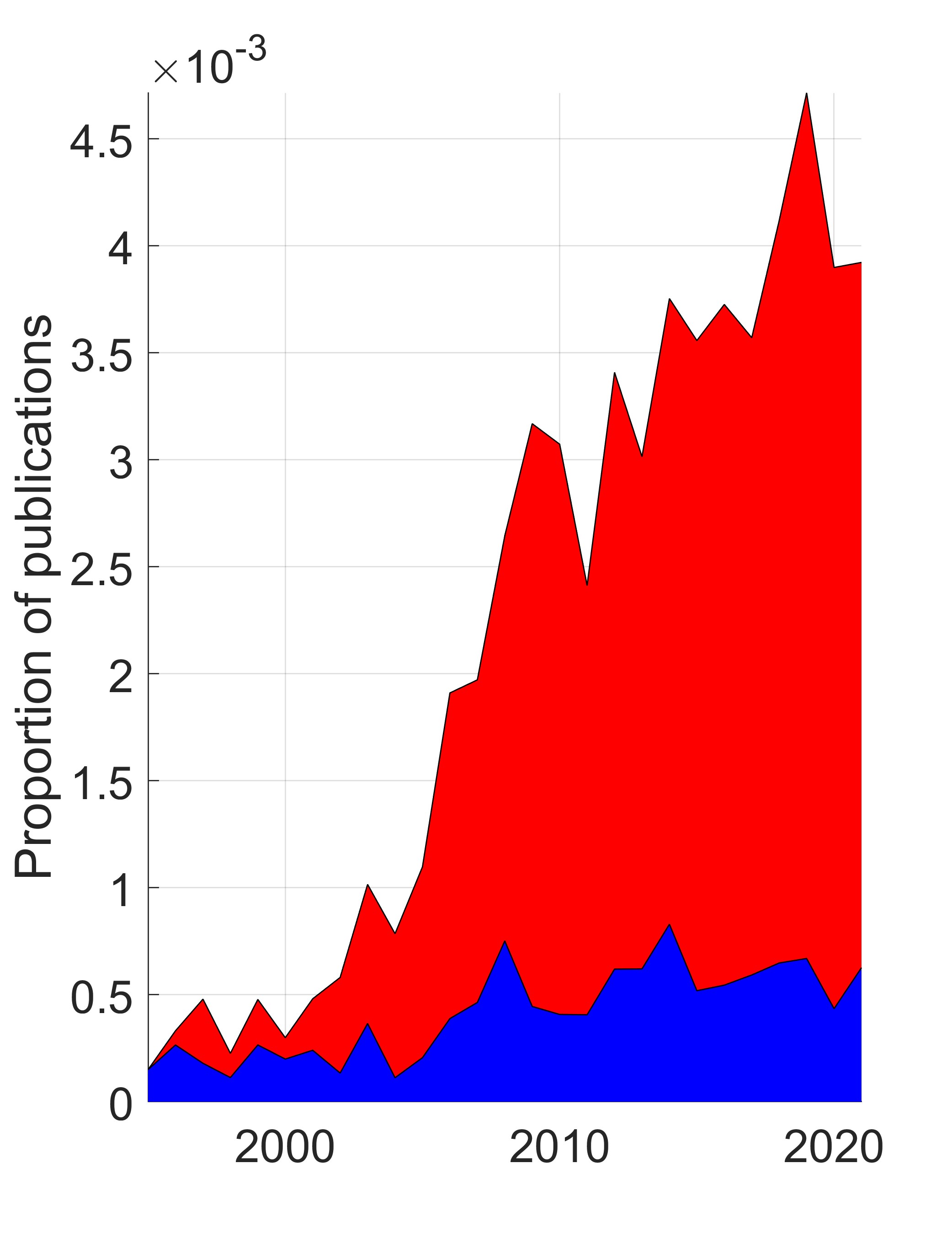}
\\
(a) Original count&(b) Scaled on&(c) Scaled on\\
&``machine learning''&``clustering''
\end{tabular}
    \caption{Number of publications retrieved from Web-of-Science a of 10th May 2022 using two keyword combinations: (1) ``constrained clustering'' (CC), and (2) ``constrained clustering'' OR (``semi-supervised'' AND clustering) (CC or  SS).}
    \label{all_publlication_count}
\end{figure}

To create an overview of the importance and development of constrained clustering, we carried out publication search using the Web-of-Science platform \url{https://www.webofscience.com/}. The reported results are valid as of 20th May 2022. However, the number of publications in the years before 2021 is unlikely to change if a search is carried out at a later date. We tried two combinations of keywords: (1) ``constrained clustering'' (CC), and (2) ``constrained clustering'' OR (``semi-supervised'' AND clustering) (CC or SS). The quotation marks indicate that the word combination was kept intact. The search was on `Topic', which includes title, abstract and keywords. Figure~\ref{all_publlication_count} plots the number of publications over the years from 1995 to 2021. 

The total number of publications on both CC and CC or SS (subplot (a)) is rising along time but this could be due to the overall trend of rising number of publications in the world. To correct for this effect, we also recovered the number of publications on ``machine learning'' (ML) and on ``clustering'' (C). Subplots (b) and (c) show the scaled number of publications obtained by dividing the original counts in subplot (a) to the respective counts for ML and C. Interestingly, according to the ML scaling, the interest in constraint and semi-supervised clustering seems to decline, especially from 2016 onward. This is likely a result of the ongoing boom of publications on deep learning, which overpower other branches of machine learning. Indeed, in terms of clustering only (subplot (c)), constrained and supervised clustering both seem to have a steady upward trend. Arguably, the term `semi-supervised clustering' is more popular and the publication proportion is increasing more notably compared to that for `constrained clustering'.

In view of the increasing publication counts, surveys and systematisation of the area of semi-supervised and constrained clustering would be beneficial to the research community.

\begin{figure}[htb]
    \centering
\begin{tabular}{cc}
\begin{minipage}{0.56\textwidth}
\includegraphics[width = 1\textwidth]{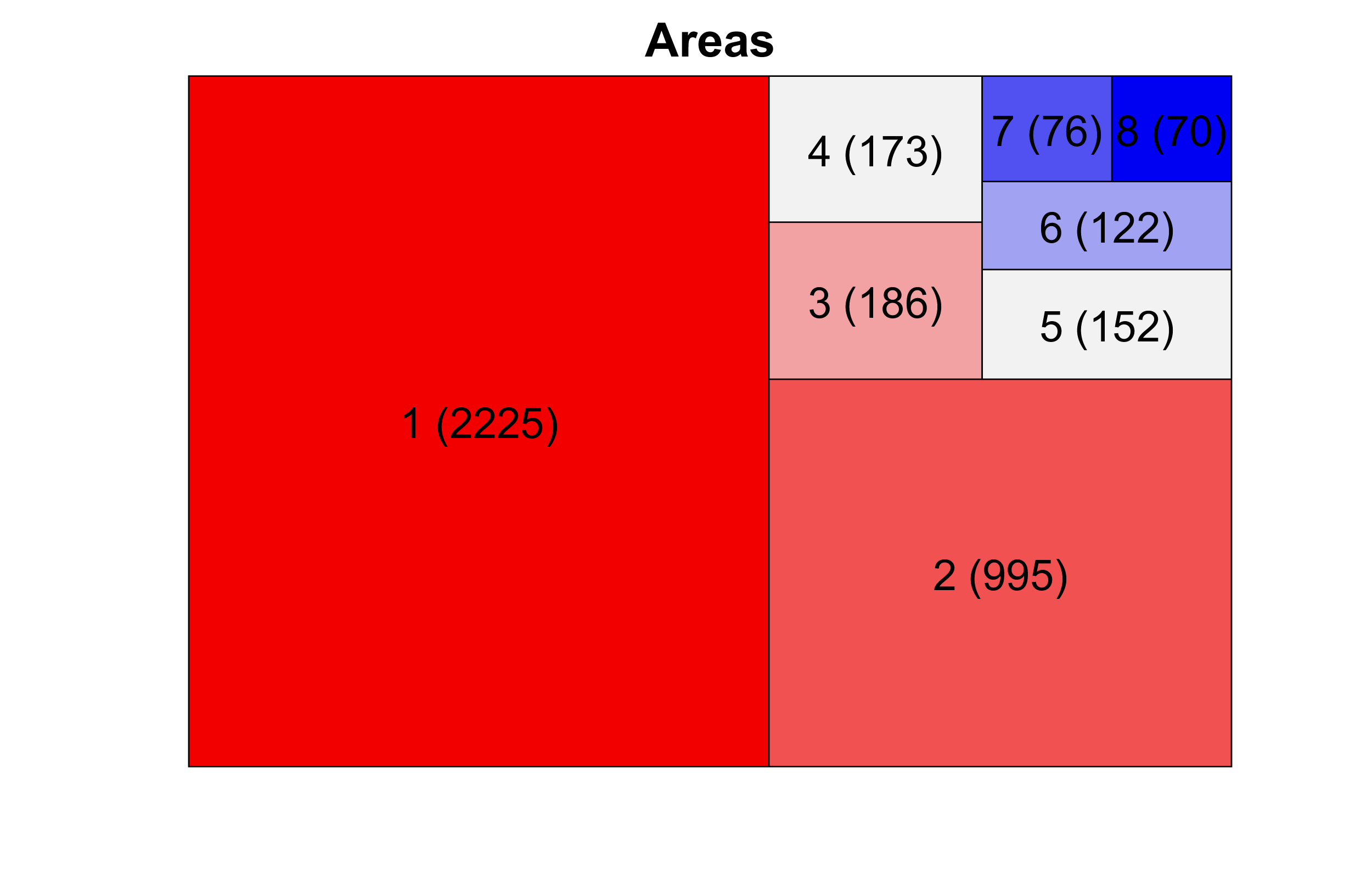}
\end{minipage}
&
\begin{minipage}{0.4\textwidth}
1. Computer Science\\
2. Engineering\\
3. Telecommunications\\
4. Mathematics\\
5. Imaging Science\\
6. Automation Control Systems\\
7. Computational Biology\\
8. Operations Research\\
\end{minipage}
\\
\end{tabular}
    \caption{Top 8 areas of research according to Web-of-Science as of 20th May 2022. All fields were searched for query ``constrained clustering'' OR (semi-supervised AND clustering).}
    \label{fig:area}
\end{figure}

Figure~\ref{fig:area} displays a tree map diagram of the top eight areas represented within the 2,981 documents retrieved with the query ``constrained clustering'' OR (semi-supervised AND clustering) from Web of Science. The area distribution was obtained from Web-of-Science.

The diagram shows that the overwhelming majority of the publications are on the technology side although some application areas are also present (telecommunications, automation control systems and computational biology). 

\begin{figure}[htb]
    \centering
\begin{tabular}{rl}

\begin{minipage}{0.5\textwidth}
\flushright
k-means or kmeans\\
must-link or pairwise or pair-wise\\
fuzzy\\
spectral\\
kernel\\
density\\
ensemble\\
``deep learning'' or ``deep neural''\\
hierarchical\\
``active learning''\\
``non-negative matrix factorization''\\
evolutionary\\
\end{minipage}&
\begin{minipage}{0.5\textwidth}
\includegraphics[width = 1\textwidth]{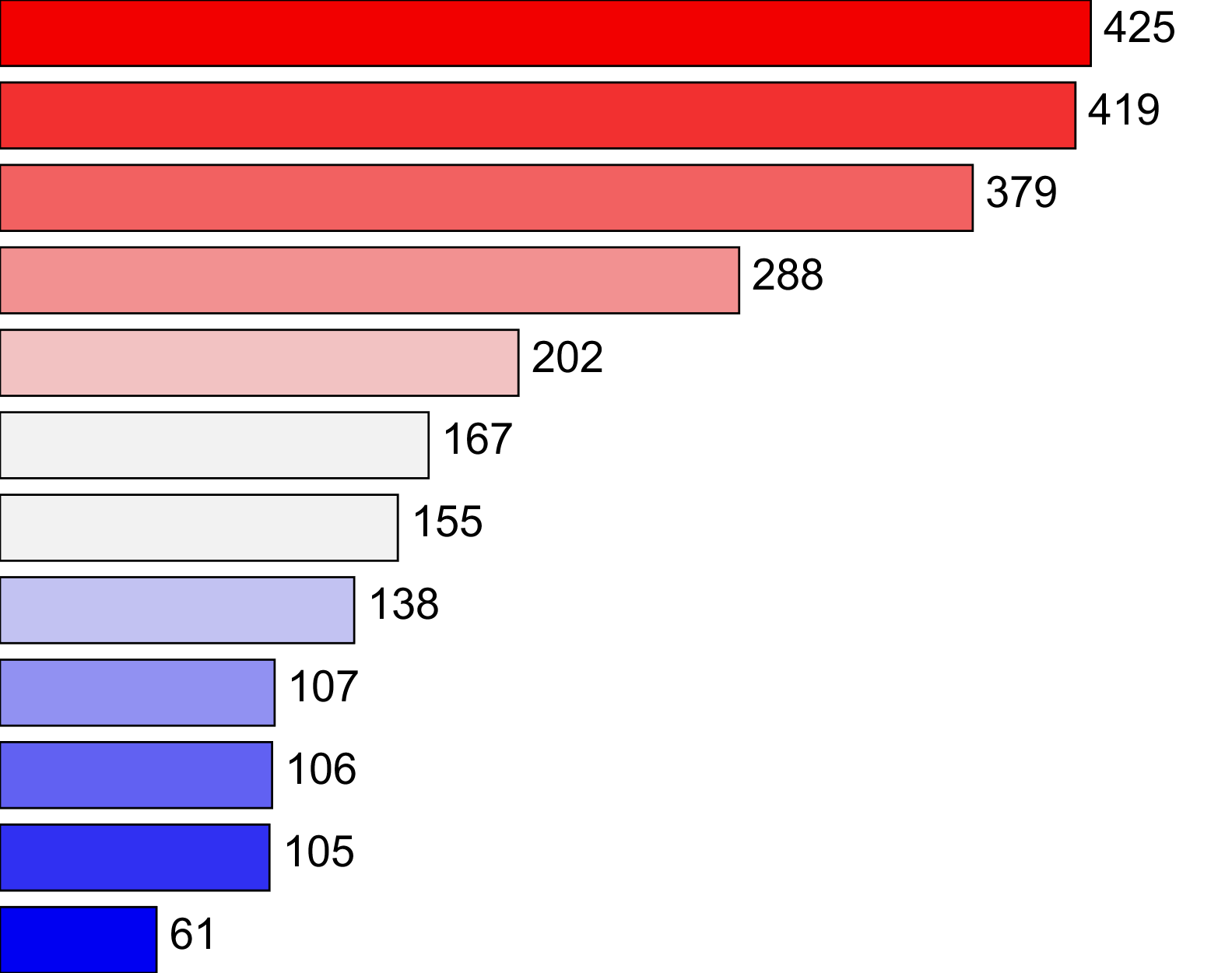}
\end{minipage}
\\
\end{tabular}
    \caption{Selected keywords arranged by frequency of occurrence within the as 2,981 documents retrieved from Web of Science on the 20th May 2022.}
    \label{fig:key}
\end{figure}

A keyword search was carried out of the retrieved papers (again, all fields were searched) using a set of relevant terms for constrained clustering. We chose not to include terms such as `graph' or `optimisation' which are relevant but too generic, and the high frequency will overshadow more focused constrained clustering terms. The results are shown in Figure~\ref{fig:key}.

As also identified by other surveys, k-means is the undisputed leader in the area of constrained clustering. `Pairwise' and `spectral' were also expected at the top of the diagram along with `kernel' and `density'. Interestingly, `fuzzy' and `ensemble'-based approaches have been mentioned often enough to secure places ahead of `deep learning' and `active learning'. We should emphasise that we are presenting a snapshot of the area at the current moment, and the keyword frequency distribution may change in the future. Most likely, `deep learning' and keywords coming from persistent application areas will climb up the diagram.

\section{Topics}
\label{sec:topics}
We collated references from the previous surveys, from Google search and Web of Science. We browsed the references to identify research focused primarily on constrained clustering, so we cast aside papers where the main thrust was an application of where constrained clustering was a secondary topic. As a result, we offer the reader a collection of references available at \url{https://github.com/LucyKuncheva/Semi-supervised-and-Constrained-Clustering/blob/main/ConstrainedClusteringReferences.pdf}. We decided to store the reference list separately so that we have space for the bibliometric analyses. We used this collection to carry out our Pareto analysis, which we see as one of the main contributions of this paper.

Creating a taxonomy of the literature on constrained clustering is not straightforward. Below we offer a rough, non-mutually exclusive grouping, acknowledging that the topics are not the same level of the hierarchy. We were guided by the interest in a particular topic, as well as by topics identified in previous surveys~\cite{Gancarski20}.

\begin{enumerate}
\item{\em K-means variants.} 
~\cite{Wagstaff01}, \cite{Wagstaff00}, \cite{tung2001constraint}, \cite{vouros2021semi}, \cite{shukla2020semi}, \cite{khashabi2015clustering}, \cite{huang2008semi}, \cite{Basu02}, \cite{Hong08}, \cite{tan2010improved}, \cite{rutayisire2011modified}, \cite{Covoes13}, \cite{Davidson06}, \cite{Amorim12}, \cite{Ge07}, \cite{Zhigang13}, \cite{banerjee2006scalable}, \cite{bradley2000constrained}, \cite{demiriz2008using}, \cite{li2007solving}, \cite{ng2000note}, \cite{kulis2009semi}, \cite{yin2010semi}

\medskip
\item{\em Spectral clustering.}
\cite{hoi2007learning}, \cite{cucuringu2016simple}, \cite{kamvar2003spectral}, \cite{alzate2009regularized}, \cite{Wang14}, \cite{wang2010active}, \cite{wang2010flexible}, \cite{Zhi13}, \cite{anand2011graph}, \cite{lu2008constrained}, \cite{lu2010constrained}, \cite{chen2021single}, \cite{ding2013research}, \cite{zhang2005analysis}, \cite{ding2018semi}, \cite{chen2012spectral}, \cite{li2008pairwise}, \cite{li2009constrained}, \cite{rangapuram2012constrained}

\medskip
\item{\em Other clustering algorithms.} hierarchical clustering~
\cite{davidson05}, \cite{xiao2016semi}, \cite{gilpin2011incorporating}, \cite{gilpin2017flexible}, \cite{yang2020interactive}, \cite{guo2008regionalization}, 
DBSCAN 
\cite{fang2019semi}, \cite{malzer2021constraint}, 
density-based clustering 
\cite{yan2021semi}, \cite{Shental03}, \cite{lelis2009semi}, 
mean-shift 
\cite{anand2014semi},\cite{tuzel2009kernel}, 
neural network-based 
\cite{Hsu16}, 
multi-view and multi-source clustering 
\cite{whang2020mega}, \cite{bai2020semi}, \cite{Ghasemi22}, \cite{Cao15}, 
or model selection 
\cite{Pourrajabi14}

\medskip
\item{\em Cluster ensembles.} 
\cite{al2009clustering}, \cite{yang2022semi}, \cite{wang2014semi}, \cite{dong2020survey}, \cite{ienco2018semi}, \cite{lai2019adaptive}, \cite{yu2017adaptive}, \cite{tian2019stratified}, \cite{yu2016incremental}, \cite{ren2019semi}, \cite{hosseini2016ensemble}, \cite{greene2007constraint}, \cite{dong2020survey}, \cite{Nan21}, \cite{dimitriadou2002mixed}, \cite{forestier2010collaborative}, \cite{iqbal2012semi}, \cite{xiao2016semi}, \cite{yang2017cluster}, \cite{yang2012consensus}, \cite{yu2011knowledge}

\medskip
\item{\em Deep learning methods.} 
\cite{ienco2018semi}, \cite{li2020semi}, \cite{chen2021single}, \cite{smieja2020classification}, \cite{shukla2020semi}, \cite{vilhagra2020textcsn}, \cite{xian2020cyber}, \cite{ren2019semi}, \cite{Lafabregue19}, \cite{Lin20}, \cite{Zhang19}

\medskip
\item{\em Soft computing approaches.} fuzzy k-means
\cite{grira2006fuzzy}, \cite{Abin15}, \cite{lai2020semi}, \cite{fantoukh2020automatic}, \cite{liu2003evolutionary}, 
evidential k-means
\cite{antoine2021fast}, ~\cite{antoine2012cecm},   
evolutionary approaches
[116],
\cite{de2017comparison}, \cite{gonzalez2020dils}, \cite{gonzalez2021me}, \cite{gonzalez2021enhancing}, \cite{luo2021pareto}, \cite{liu2003evolutionary} 
ant colony optimisation 
~\cite{yang2015parallel}

\medskip
\item{\em Learning a distance metric.} 
\cite{tang2007enhancing},\cite{bar2003learning}, \cite{bar2005learning}, \cite{Klein02}, \cite{xing2002distance}, \cite{Cohn03}, \cite{bilenko2004integrating}, \cite{hoi2010semi}, \cite{yi2012semi}, \cite{wang2013semi}, \cite{li2020semi}, \cite{guo2021joint}, \cite{abin2020learning}, 

\medskip
\item{\em Incorporating the constraints into the criterion function.}
\cite{demiriz1999semi}, \cite{davidson2005clustering}, \cite{pelleg2007k}, \cite{Basu04}, \cite{Hiep16}, \cite{melnykov2020note}, \cite{ganji2016lagrangian}, \cite{bilenko2004integrating}, \cite{basu2004probabilistic}, \cite{kulis2013metric}, \cite{cheng2008constrained}, \cite{dao2013declarative}, \cite{Dao16}, \cite{babaki2014constrained}, \cite{chabert2017constraint}, \cite{davidson2010sat}, \cite{berg2017cost}, \cite{metivier2012constrained}, \cite{tang2019size}, \cite{mueller2010integer}, \cite{ouali2016efficiently}, \cite{Dao16}, \cite{Dao17}, \cite{lampert2018constrained}, \cite{khiari2010constraint}


\medskip
\item{\em Active learning, user interaction, incremental clustering.}
\cite{wang2010active}, \cite{Yang20interactive}, \cite{li2021learning}, \cite{li2019ascent}, \cite{fernandes2020improving}, \cite{abin2020density}, \cite{chen2020active}, \cite{awasthi2014local}, \cite{chang2016appgrouper}, \cite{coden2017method}, \cite{prabakara2013incremental}, \cite{srivastava2016clustering}, \cite{vu2017active}, \cite{yang2020interactive}

\medskip
\item{\em Applications.} 

In addition to the theoretical and algorithmic advances in constrained clustering, we came across a beautiful variety of applications, among which: analysis of RNA~\cite{Tian21}, \cite{chen2021single} gene expression data analysis~\cite{wang2014semi}, medical imaging~\cite{xia2020oriented}, EEG data analysis~\cite{du2019method}, vegetation classification~\cite{tichy2014semi}, regionalisation using spatial contiguity constraints ~\cite{brenden2008spatially}, \cite{guo2008regionalization}, \cite{kupfer2012regionalization}, \cite{patil2006spatially}, text and document clustering~\cite{Yang20interactive}, \cite{sadjadi2021two}, \cite{buatoom2020document}, \cite{vilhagra2020textcsn}, information retrieval~\cite{Zhan22}, object and face clustering in video~\cite{Wu13}, \cite{Kalogeiton20}, \cite{Cao15}, \cite{Kulshreshtha18}, \cite{Yan06}, \cite{arachchilage2020adaptive},  tracking of moving objects using radar sensors~\cite{malzer2021constraint}, time series clustering~\cite{he2019fast}, \cite{lampert2018constrained}, tourism~\cite{bernini2021spatial}, financial analysis~\cite{han2021semi}, \cite{zhang2022clustering}, defect prediction~\cite{zhang2017label}, \cite{li2014semi}, cyber security~\cite{xian2020cyber}, \cite{huda2017defending}, \cite{gu2019semi}, and  malware clustering~\cite{fang2019semi}.
\end{enumerate}

\section{Pareto analysis}
\label{sec:par}

We assume that papers published more recently and papers with a large number of citations are generally more influential. These two criteria are not likely to be satisfied by a single paper. A compromise between year of publication and number of citations should be sought. Admittedly, most recent papers would not have had enough exposure to attract a large number of citations yet, but this does not impact on their future relevance. Pareto front contains all {\em non-dominated} alternatives. A paper $x$ is in the Pareto front if there is no other paper $y$ in the collection which is better than $x$ on {\em both} criteria.

\begin{figure} [h]
    \begin{centering}
    \includegraphics[width = \textwidth]{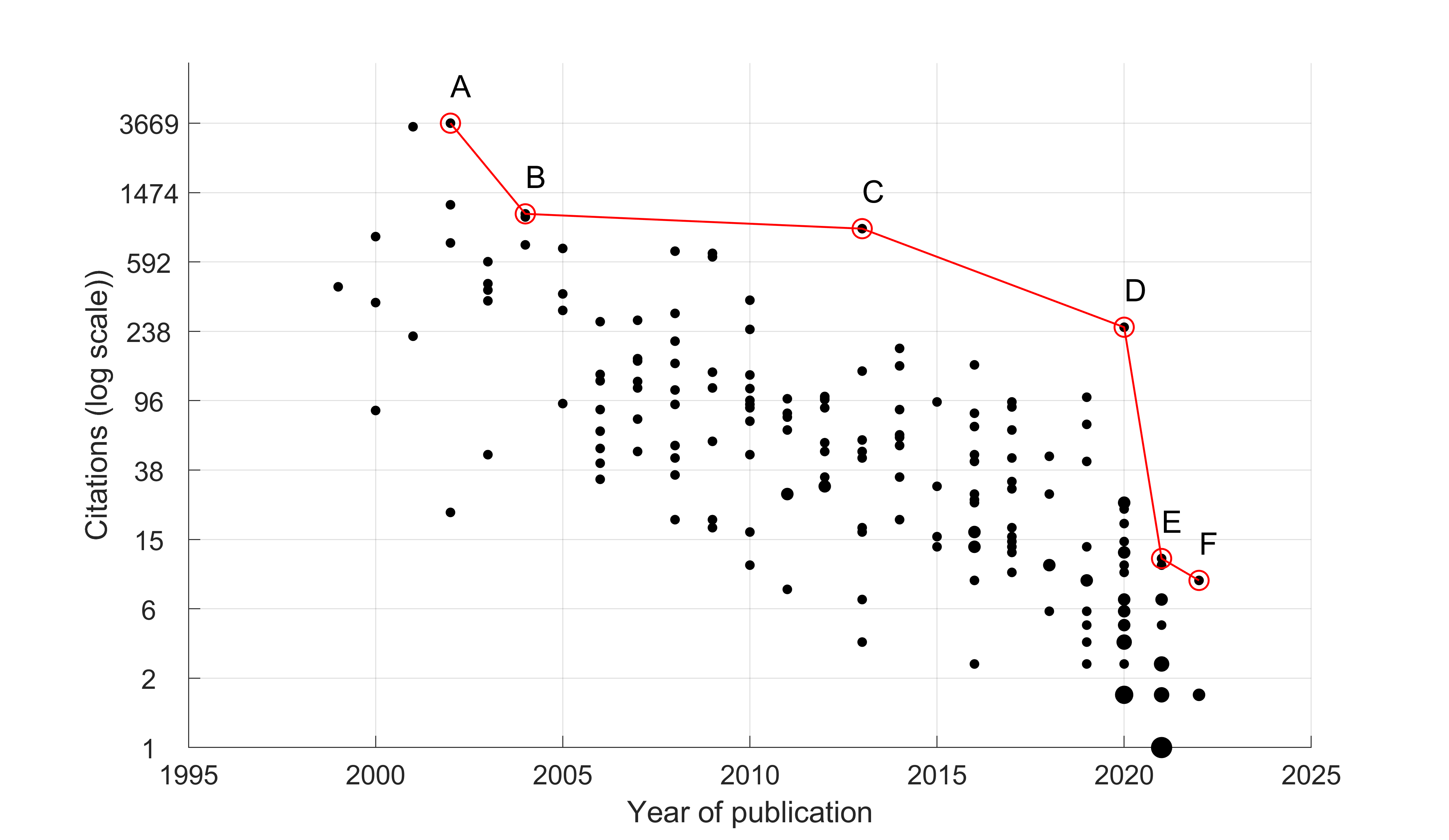}
    \end{centering}
    
\begin{itemize}
\item A: \cite{xing2002distance}, 3669 citations, Xing et al. (2002)
Distance metric learning with application to clustering with side-information
\item B: \cite{bilenko2004integrating}, 1114 citations, Bilenko et al. (2004)
Integrating constraints and metric learning in semi-supervised clustering
\item C: \cite{kulis2013metric}, 917 citations, Kulis et al. (2013)
Metric learning: A survey
\item D: \cite{dong2020survey}, 251 citations, Dong et al. (2020)
A survey on ensemble learning
\item E: \cite{Tian21}, 12 citations, Tian et al. (2021)
Model-based deep embedding for constrained clustering analysis of single cell {RNA}-seq data
\item F: \cite{Zhan22}, 9 citations, Zhan et al. (2022)
Learning discrete representations via constrained clustering for effective and efficient dense retrieval

\end{itemize}
    \caption{Pareto front of all reviewed publications (as of 20th May 2022).}
    \label{fig:paretoall}
\end{figure}

Figure~\ref{fig:paretoall} shows a scatterplot of the papers covered in this survey in the space (year-of-publication, $\log(K)$), where $K$ is the number of citations according to Google Scholar, as of 20th May 2022. The Pareto front is marked with a solid red line, and the papers in it are listed under the figure. Some dots are larger than others to indicate that there are more than one paper published in the same year, with the same number of citations.

It is interesting to notice that the earlier papers are mostly on learning a distance metric, more recent ones are on clustering methodologies (ensembles and deep learning), and the most recent two papers in the Pareto front are application-orientated.

\begin{figure}
    \centering
    \includegraphics[width = 1\textwidth]
    {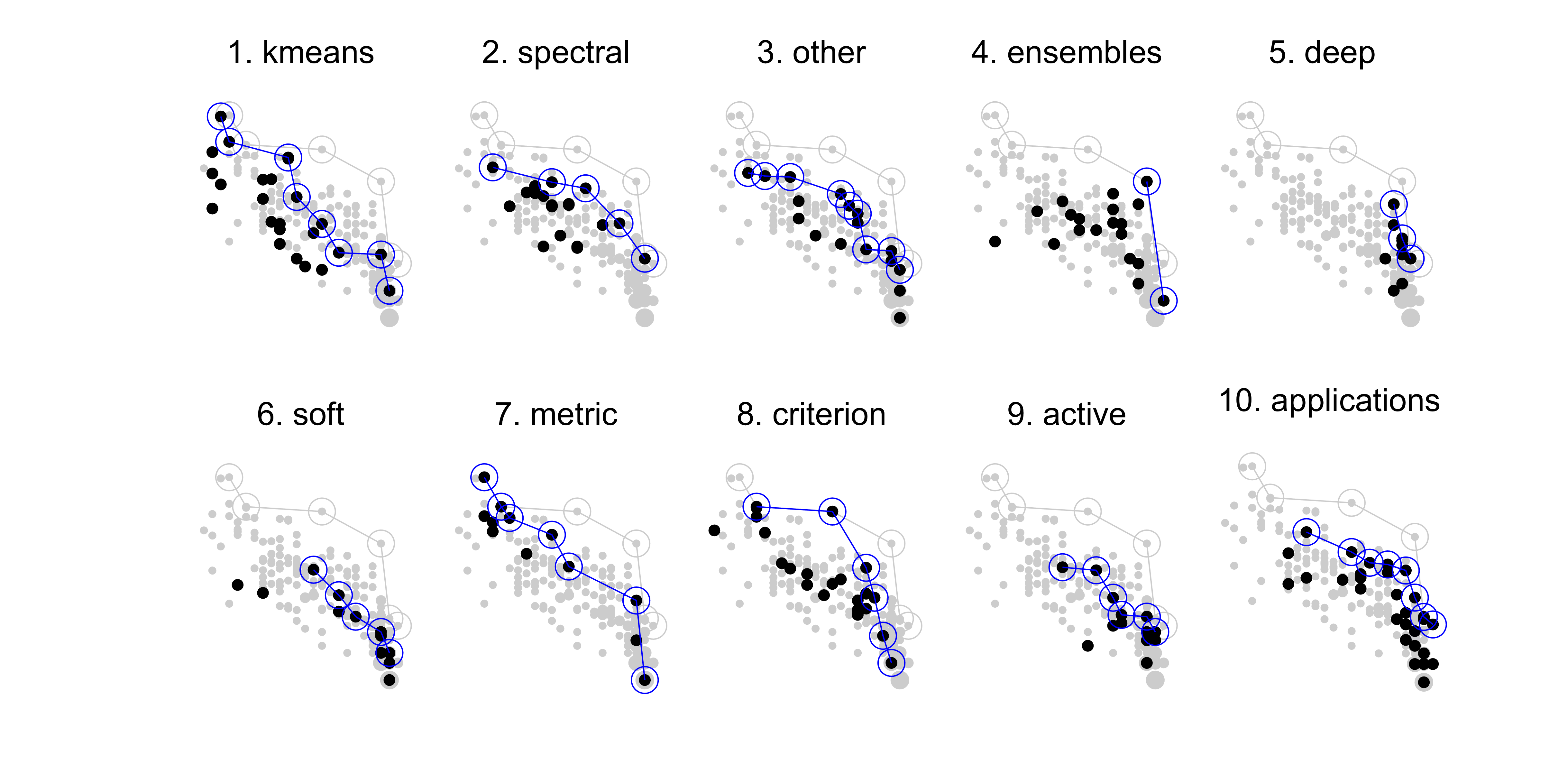}
    \caption{Pareto front for the ten topics.}
    \label{fig:pakm}
\end{figure}

\begin{figure}
    \centering
    \includegraphics[width = 1\textwidth]
    {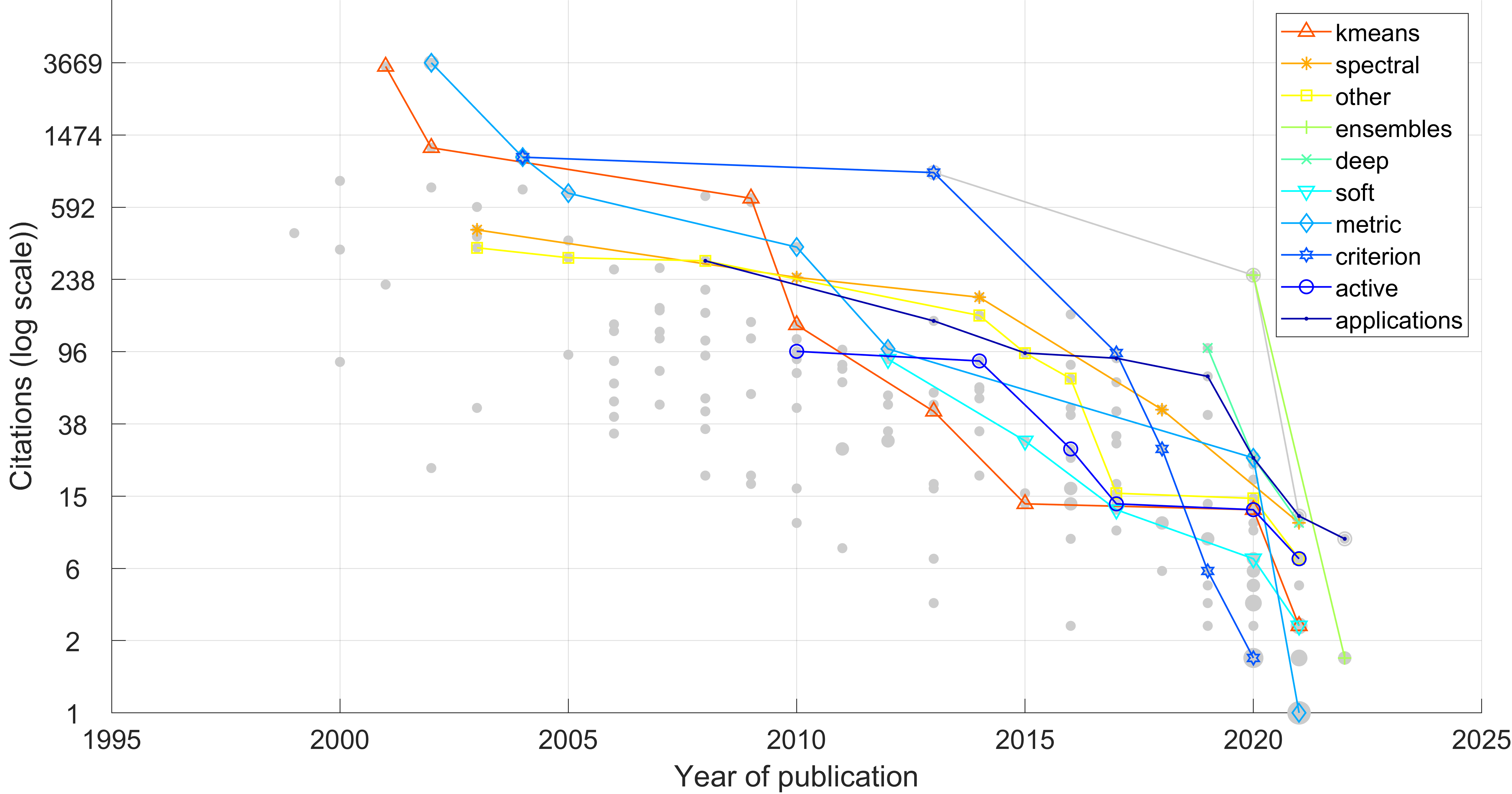}
    \caption{Pareto front for the ten topics on the same graph.}
    \label{fig:topto}
\end{figure}

Figure~\ref{fig:pakm} shows the Pareto front for the ten individual topics and Figure~\ref{fig:topto} shows all the 10 Pareto Fronts together.

The plots indicate that more recent interest in constrained clustering is focused on applications, deep learning, soft computing, and active learning. On the other hand, developments on k-means and spectral clustering were less represented in the more recent literature. Compared to the overall Pareto front, the ensemble methods, deep learning, and the applications seem to be dominating the rest of the topics in terms of recent citation counts. Category 3, `other methods' seems to be enjoying a healthy interest all throughout the years.

\section{Available software}
\label{sec:sof}

Table~\ref{tab:as} shows a list of software for constrained clustering which we found. Our list is by no means exhaustive.

\renewcommand{\arraystretch}{1.3}
\setlength{\tabcolsep}{8pt}
\begin{table}[p]
\caption{Available software for constrained clustering}
\label{tab:as}
\centering 
\begin{tabular}{p{4.4cm}p{1cm}p{6cm}} 
Algorithm& Year& Link\\
\hline
Auto-tuning spectral clustering&2022&
\url{https://github.com/tango4j/Auto-Tuning-Spectral-Clustering}\\\hline
Binary optimisation constrained k-means (BCKM)&2019&
\url{https://github.com/intellhave/BCKM}\\\hline
Cluster fractional allocation matrix (CFAM)&2020&
\url{https://github.com/dung321046/ConstrainedClusteringViaPostProcessing}\\\hline
Constrained deep adaptive clustering (CDAP)&2020&
\url{https://github.com/thuiar/CDAC-plus}\\\hline
Constrained graph clustering&2017&
\url{https://github.com/Behrouz-Babaki/Pigeon}\\\hline
Constrained K-means&2017&
\url{https://github.com/NestorRV/constrained_kmeans}\\\hline
Constrained online face clustering (COFC)&2018&
\url{https://github.com/ankuPRK/COFC}\\\hline
Constraint satisfaction clustering&2022&
\url{https://github.com/autonlab/constrained-clustering}\\\hline
COP/PC K-means&2019&
\url{https://github.com/ashkanmradi/constrained-k-means}\\\hline
COPK-means&2017&
\url{https://github.com/Behrouz-Babaki/COP-Kmeans}\\\hline
DCDS&2019&
\url{https://github.com/leule/DCDS}\\\hline
Deep Constrained Clustering&2020&
\url{https://github.com/blueocean92/deep_constrained_clustering}\\\hline
Lpbox-ADMM&2021&
\url{https://github.com/wubaoyuan/Lpbox-ADMM}\\\hline
MinSizeK-means&2021&
\url{https://github.com/Behrouz-Babaki/MinSizeKmeans}\\\hline
MIPK-means&2017&
\url{https://github.com/Behrouz-Babaki/MIPKmeans}\\\hline
PC-SOS-SDP&2021&
\url{https://github.com/antoniosudoso/pc-sos-sdp}\\\hline
repCONC&2022&
\url{https://github.com/jingtaozhan/repconc}\\\hline
Spectral clustering with fair constraints&2019&
\url{https://github.com/matthklein/fair_spectral_clustering}\\\hline
SpectralNet&2020&
\url{https://github.com/KlugerLab/SpectralNet}\\\hline
\end{tabular}
\end{table}

\section{Experimental studies}
\label{sec:exp}

Our analysis of the literature revealed a marked lack of large comparative studies. We gathered and summarised the experiments from the cited works which reported an experiment; a total of 76 publications. For each experiment, we collated the list of data sets used, the algorithms compared, and the evaluation metrics.

\subsection{Datasets}
\begin{figure}[htb]
    \centering
\begin{tabular}{rl}

\begin{minipage}{0.65\textwidth}
\flushright
Iris\\
Wine\\
Glass\\
Ionosphere\\
MNIST\\
Ecoli\\
Novel Dataset, Breast Cancer\\
Letters, Soybean\\
Pima, 20-Newsgroups, Vehicle, Digits, Balance, Heart\\
\end{minipage}&
\begin{minipage}{0.3\textwidth}
\includegraphics[width = 1\textwidth]{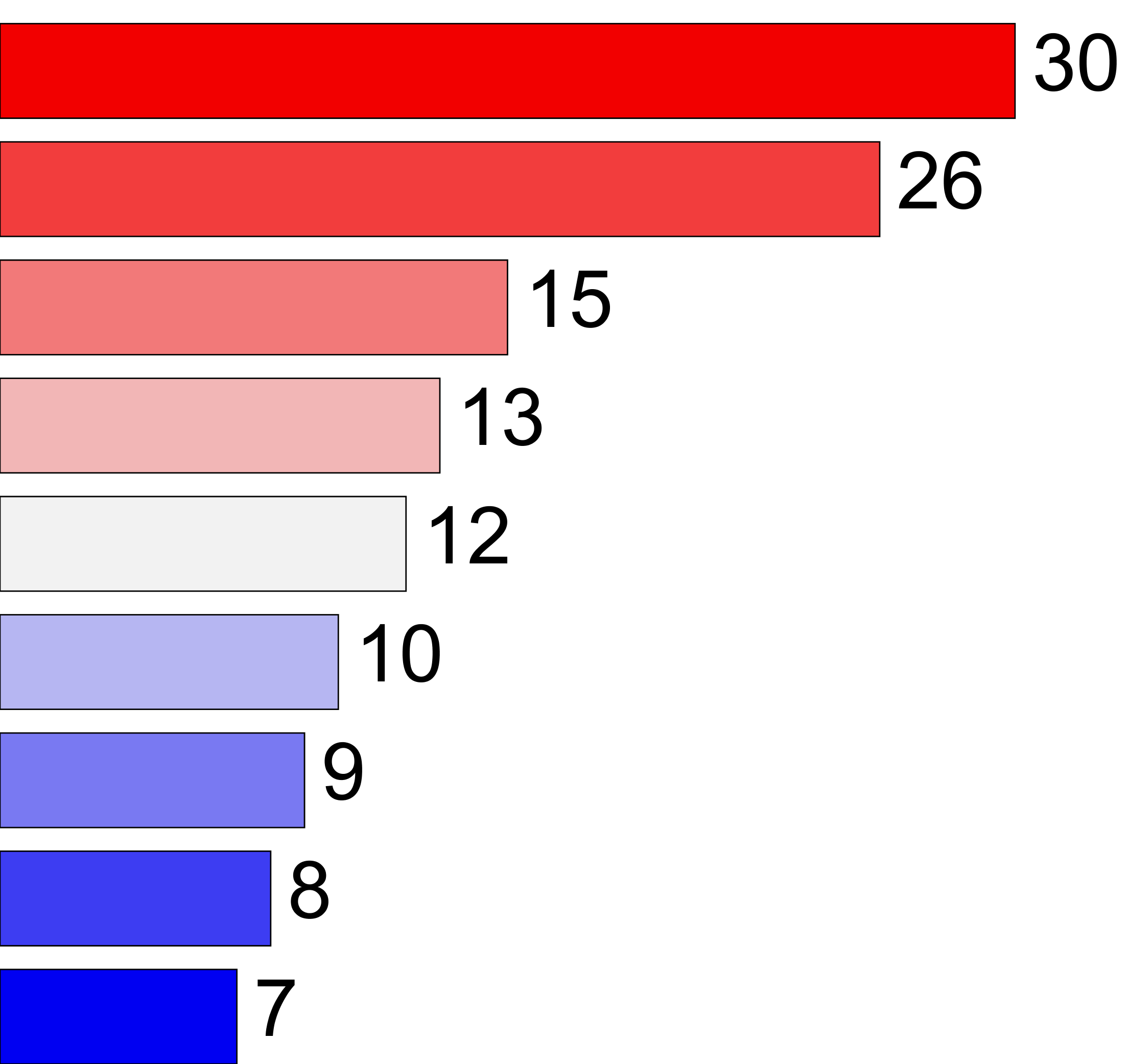}
\end{minipage}
\\
\end{tabular}
    \caption{Histogram of datasets that appeared more than six times in the constrained clustering literature}
    \label{fig:exp_ds}
\end{figure}

A total of 245 datasets were identified. Out of these, 179 were used only once, and 30 were used twice, demonstrating the deficiency in comparative studies. Figure~\ref{fig:exp_ds} displays a histogram of the datasets that have been used more than six times throughout the experimental studies. The four most commonly used datsets were: Iris, Wine, Glass, and Ionosphere from the UCI Machine Learning Repository \url{https://archive.ics.uci.edu/ml/}.

\begin{figure}[htb]
    \centering
    \includegraphics[width = 0.8\textwidth]{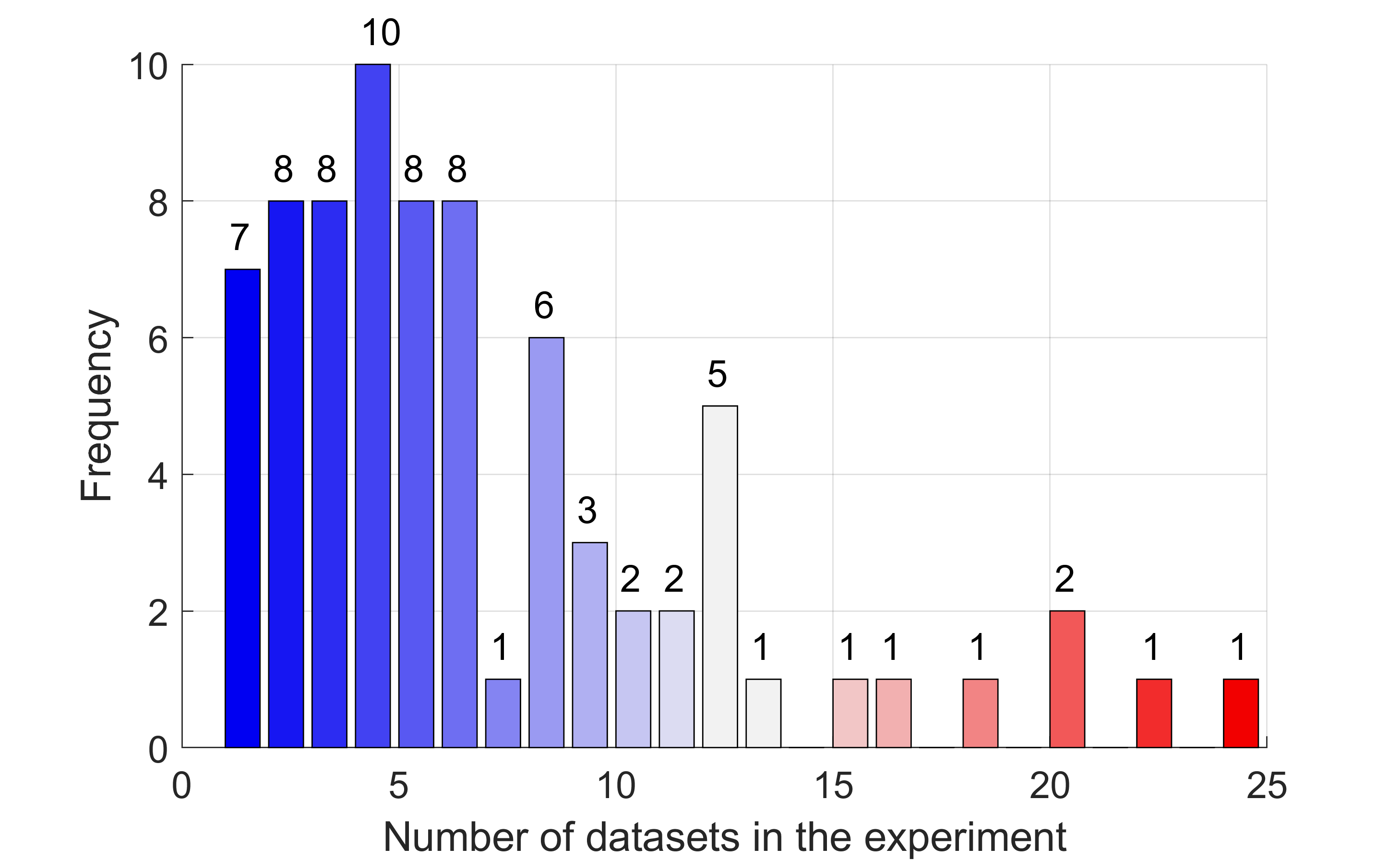}
    \caption{Histogram of the number of datasets used in the experiments.}
    \label{fig:exp_noDs}
\end{figure}

To complement this result, Figure~\ref{fig:exp_noDs} shows a histogram of the {\em number of datasets} used in the experiments reported in the literature. It can be seen that only a few works use more than 10 datasets, which, again, points at the lack of adequate large-scale comparisons.

\subsection{Algorithms}

\begin{figure}[htb]
    \centering
\begin{tabular}{rl}

\begin{minipage}{0.45\textwidth}
\flushright
KMeans\\
COP-KMeans\\
MPC-KMeans\\
PC-KMeans\\
Constrained-KMeans\\
LCVQE, E$^2$CP, DBSCAN\\
\end{minipage}&
\begin{minipage}{0.3\textwidth}
\includegraphics[height = 2.2 cm]{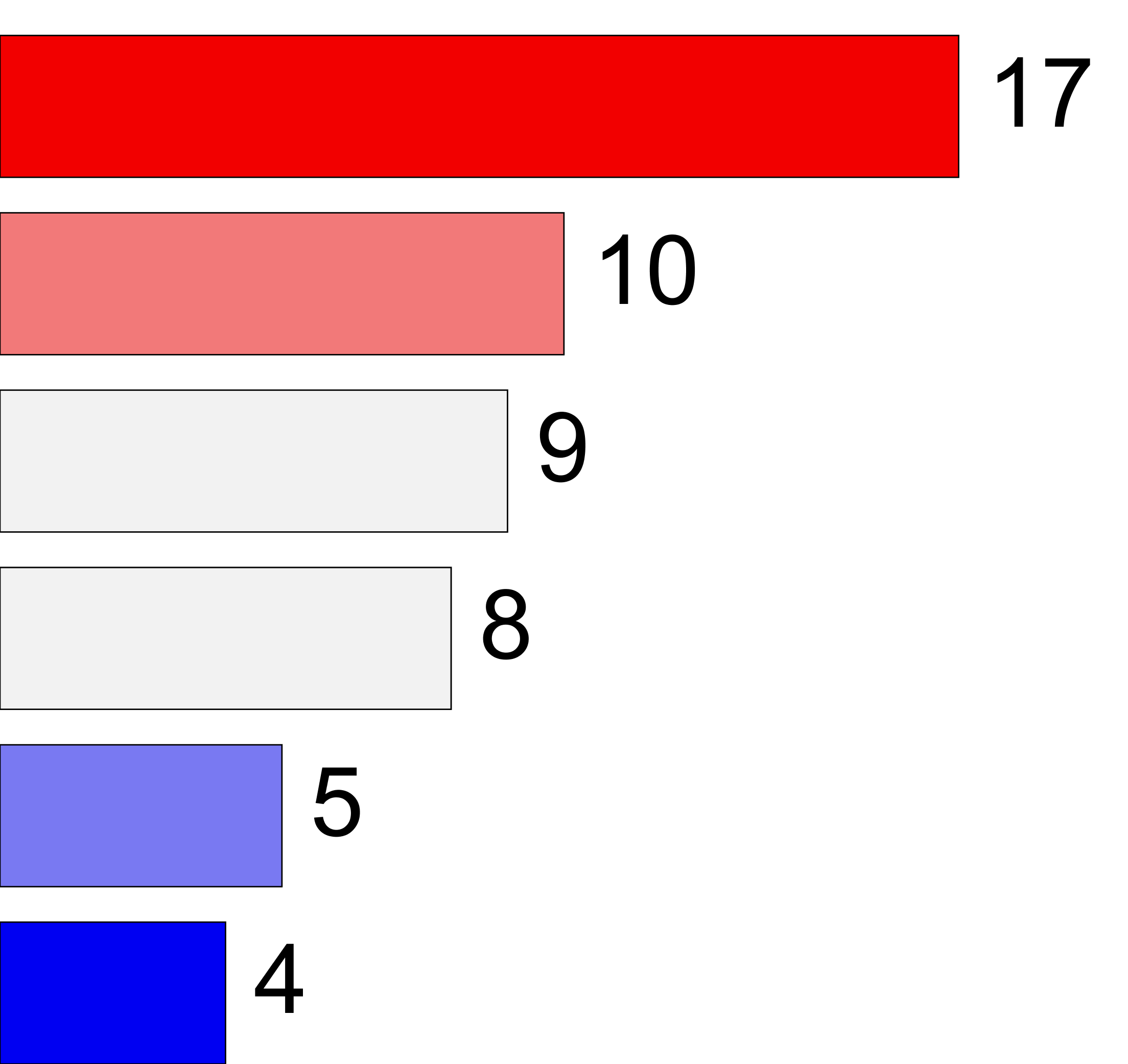}
\\\end{minipage}
\\
\end{tabular}
    \caption{Constrained clustering algorithms most used in experiments to compare against.}
    \label{fig:exp_alg}
\end{figure}

Figure~\ref{fig:exp_alg} displays the algorithms that are used for comparison within the experiments. The most commonly used algorithms were: K-Means, COP-KMeans, and MPC-KMeans.

\begin{figure}[htb]
    \centering
    \includegraphics[width = 0.7\textwidth]{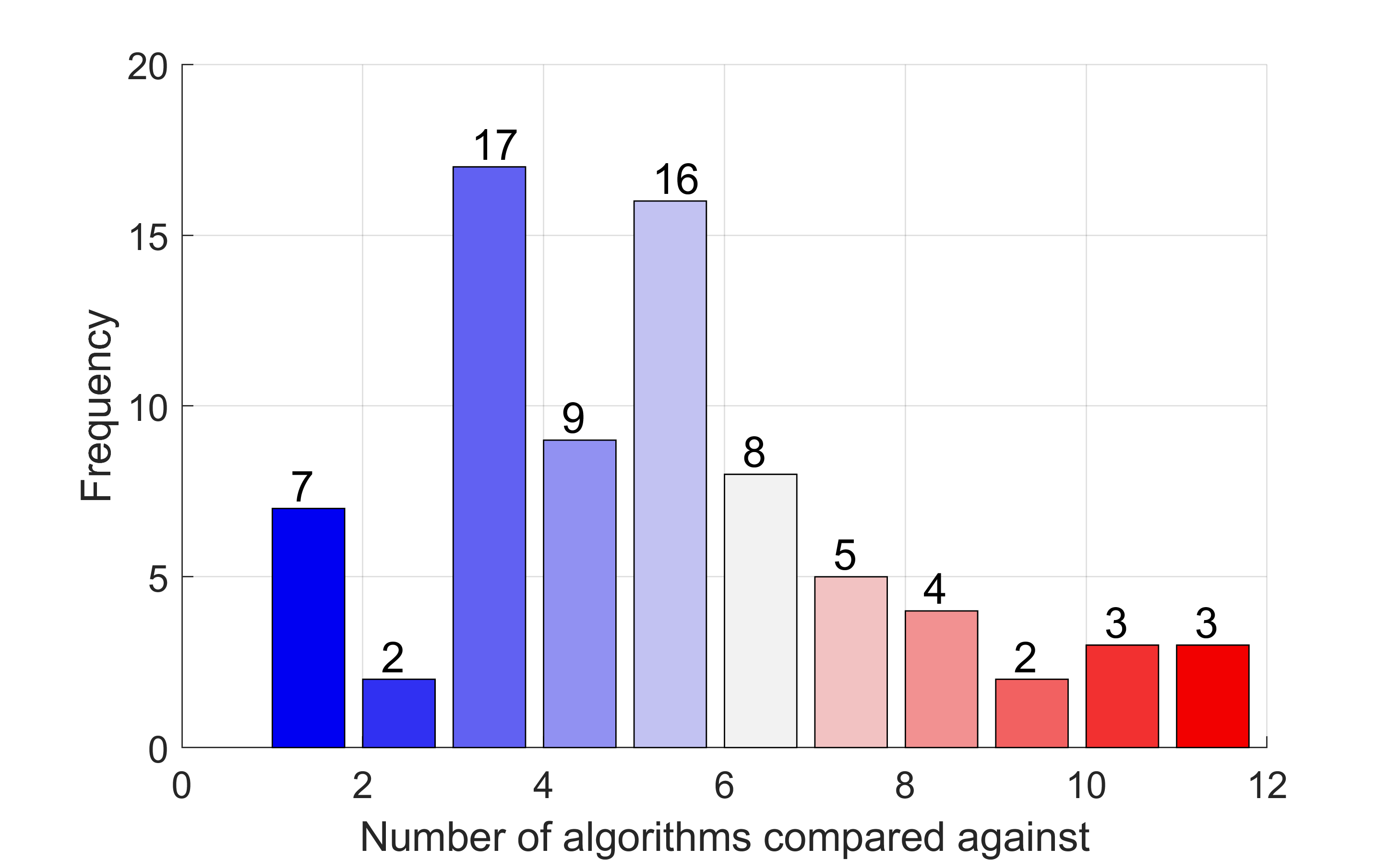}
    \caption{Histogram of the number of datasets used in the experiments.}
    \label{fig:exp_noAs}
\end{figure}

Figure~\ref{fig:exp_noAs} shows a histogram of the {\em number of algorithms compared against} used in the experiments. Only a handful of papers report comparisons between above 10 algorithms. 

\subsection{Evaluation Metrics}
\begin{figure}[htb]
    \centering
\begin{tabular}{rl}
\begin{minipage}{0.4\textwidth}
\flushright
Normalised Mutual Information\\
Clustering Accuracy\\
Adjusted Rand Index\\
F-Measure\\
Cluster Purity\\
Rand Index\\
\end{minipage}&
\begin{minipage}{0.2\textwidth}
\includegraphics[width = 1\textwidth]{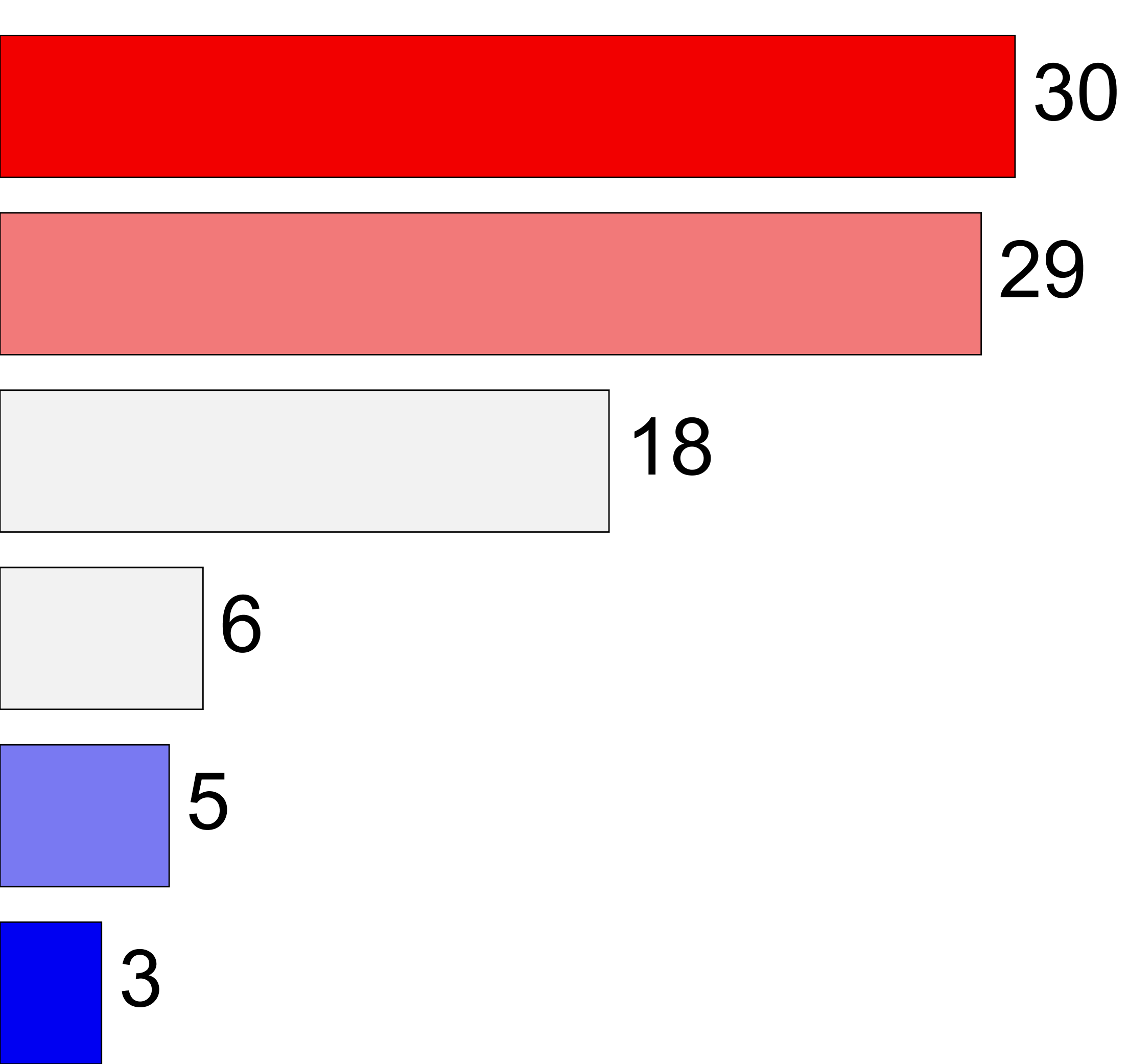}
\end{minipage}
\\
\end{tabular}
    \caption{Histogram of the evaluation metrics most often used in the experimental comparisons.}
    \label{fig:exp_em}
\end{figure}

Figure-\ref{fig:exp_em} shows the most popular metrics used to evaluate and compare constrained clustering algorithms. All there metrics rely an an external {\em labelled} dataset, which reinforces the message that, in absence of a better gauging criterion, the fundamentally flawed approach of comparing partition labels with pre-assigned labels is most often applied.

\section{Conclusion}
This work presents a bibliographic snapshot of the work on constrained clustering, as of 20 May 2022. The main sources for our analyses were Google Scholar and Web-of-Science. We did systematic keyword search in Web-of-Science, and sources citation counts from Google Scholar. 

File \verb!ConstrainedClusteringReferences.pdf!, available at \url{https://github.com/LucyKuncheva/Semi-supervised-and-Constrained-Clustering/blob/main/ConstrainedClusteringReferences.pdf}, gives a full list of the papers cited here and included in the Pareto analyses and the experimental analyses.

The main finding of our work is a notable lack of extensive experimental comparisons between methods for constrained clustering. Such comparisons would be a useful tool for present and future researchers. 

Shared software is the way forward in order to unify the approaches and facilitate comparisons. The visibility of research works which share their software is much better. 

Our Pareto analysis demonstrated that application-orientated papers will likely dominate the future development of the area, where generic methodology and algorithms will make space for more idiosyncratic ones. In terms of methodology, ensemble learning, deep learning and active/incremental learning appear to be still at the forefront. 

\section*{Acknowledgment}
This work is supported by the UKRI Centre for Doctoral Training in Artificial Intelligence, Machine Learning and Advanced Computing (AIMLAC), funded by grant EP/S023992/1.


\end{document}